\documentclass[10pt,twocolumn,letterpaper]{article}

\usepackage{iccv}      % To produce the REVIEW version
\definecolor{iccvblue}{rgb}{0.21,0.49,0.74}
\usepackage[pagebackref,breaklinks,colorlinks,allcolors=iccvblue]{hyperref}
\usepackage{multirow}
\usepackage{adjustbox} 
\usepackage{lscape}
\usepackage{booktabs}
\usepackage{caption}
\usepackage{colortbl}
\usepackage{xcolor}
\def\paperID{****} % *** Enter the Paper ID here
\def\confName{ICCV}
\def\confYear{2025}

\title{ATSTrack: Enhancing Visual-Language Tracking by Aligning Temporal and Spatial Scales}

\author{
    Yihao Zhen$^{1}$, Qiang Wang$^{2}$, Yu Qiao$^{3}$, Liangqiong Qu$^{4}$, Huijie Fan$^{1}$\thanks{Corresponding author} \\
    $^{1}$Shenyang Institute of Automation, CAS \\
    $^{2}$School of Information Engineering, Shenyang University \\
    $^{2}$School of Software, Shandong University \\
    $^{2}$The University of Hong Kong\\
    {\tt\small  }
}
\begin{document}
\maketitle
\begin{abstract}
     A main challenge of Visual-Language Tracking (VLT) is the misalignment between visual inputs and language descriptions caused by target movement. Previous trackers have explored many effective methods to preserve more aligned features. However, we have found that they overlooked the inherent differences in the temporal and spatial scale of information between visual and language features, which ultimately hinders their capability. To address this issue, we propose a novel visual-language tracker that enhances the effect of feature modification by \textbf{A}ligning \textbf{T}emporal and \textbf{S}patial scale of different input components, named as \textbf{ATSTrack}. Specifically, we decompose each language description into four phrases with different attributes based on their temporal and spatial correspondence with visual inputs, and modify their features in a fine-grained manner. Moreover, we introduce a Visual-Language token that comprises modified linguistic information from the previous frame to guide the model to extract visual features that are more relevant to language description, thereby  reducing the impact caused by the differences in spatial scale. Experimental results show that our proposed ATSTrack achieves a performance comparable to existing methods. Our code will be released.
\end{abstract}    
\section{Introduction}

Vision-Language tracking aims to track targets based on initial bounding boxes and additional natural language descriptions. This approach could overcome the limitations of relying solely on visual modalities and thus improve the tracking performance by leveraging high-level semantic information in language descriptions~\cite{mgit,dtllm,Li2024BeyondMS}. 

\begin{figure}[t]
    \centering % 使用\centering来居中图片
    \includegraphics[width=0.99\linewidth]{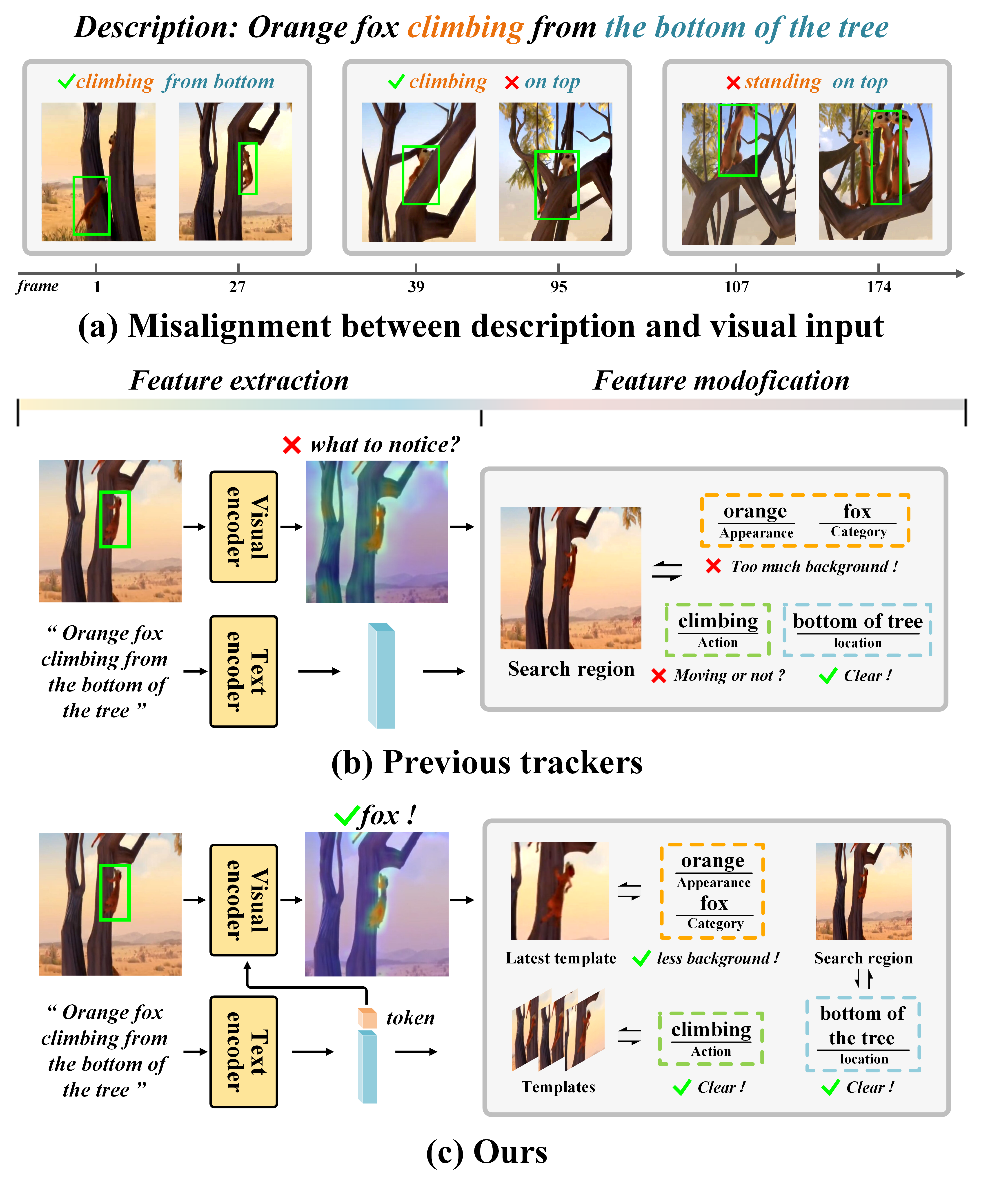} 
   \caption{Comparison with other Visual-Language trackers. (a) The mismatch between language descriptions and visual inputs. (b) Paradigm of previous trackers. (c) We utilize a token containing linguistic information to guide the extraction of visual features, and propose a fine-grained modulation module to modify language features.}
    \label{fig:shoutu}
\end{figure}

A main challenge of Visual-Language tracking is the misalignment between visual inputs and language descriptions~\cite{querynlt,JointNLT}. 
Specifically, existing language descriptions are typically a description of the target's state in the first frame or a summary over a period of time. As the target moves, it may undergo deformation or changes in action and become inconsistent with the language description, leading to a misalignment between visual and language features. As illustrated in \cref{fig:shoutu}(a), the target's action and location changed from ``climbing the tower" to ``squatting on the tower", and finally to ``flying in the air". 
Regarding this issue, it is crucial to modify language features in order to filter out the information that does not align with the current state of target. Despite some effective feature modification methods have been explored by previous visual-language trackers~\cite{citetracker,Decouple,uvl,querynlt,JointNLT,Sun2024ChatTrackerEV}, we have found that these methods overlooked the inherent differences in the temporal and spatial scales of information contained in different parts of visual and language features\cite{multi-scale,Chen2023OSTRT}, and fail to achieve the optimal modification effect.

Specifically, the description of the target itself typically corresponds only to a small portion of the image and covers a limited spatial scale compared to visual input. The action of target could encompass its states over a period of time (\eg dancing,playing) and contains more temporal information compared to traditional search-template image pairs. As illustrated in \cref{fig:shoutu} (b), previous trackers use all visual and language features as two entireties during modification, which suffers inevitable interference caused by temporal and spatial differences. For example, when using visual feature to modify the description about the target appearance, excessive background information may introduce interference.  
 To address this issue, we propose a fine-grained visual-language interaction strategy to enhance the effect of language feature modification. We replace the single template used in previous trackers with a template  sequence to incorporate more temporal visual information, and decompose language descriptions into phrases with different attributes based on their temporal and spatial correspondence with different visual inputs. Features of each attribute are then refined with the corresponding visual inputs and in different manners through a \textbf{Fine-Grained Modification(FGM)} module.

  Another problem caused by the spatial scale difference arises during the feature extraction. As mentioned above, the spatial scale of visual input is usually larger than language description. In previous trackers, visual features are extracted independently without the involvement of linguistic information, which can cause visual backbone to pay unnecessary attention to those irrelevant visual details (\eg irrelevant objects, background), while neglecting features that are related to the language description. Even if the model pays sufficient attention to the target through the interaction with the template, the focus of the features it extracts (\eg texture, edges) may still diverge from the language description (\eg color, action). To address this issue, we introduce a \textbf{Visual-Language token (VL token)} that incorporates both modified linguistic information and propagates it to the visual backbone in the following frame. 
 In such a way, the model can extract visual features that are more relevant to language descriptions with the guidance of linguistic information. 

 Our main contributions are summarized as follows:
\begin{itemize}
\item We propose ATSTrack, a novel Visual-Language Tracking framework, which could enhance the effect of feature modification by aligning temporal and spatial scale of different input components.

\item We address the interference caused by the temporal and spatial misalignment between visual and language features with a Fine-Grained Modulation module, and enhance the cross-modality correlation by using a Visual-Language token that incorporates linguistic information to guide the extraction of visual features.

\item The proposed ATSTrack outperforms state-of-the-art Vision-Language trackers on three tracking datasets. We conducted extensive experiments including ablation studies to demonstrate the effectiveness of the proposed framework and each module.
\end{itemize}
 
\begin{figure*}[t]
    \centering % 使用\centering来居中图片
    \includegraphics[width=0.9\linewidth]{
    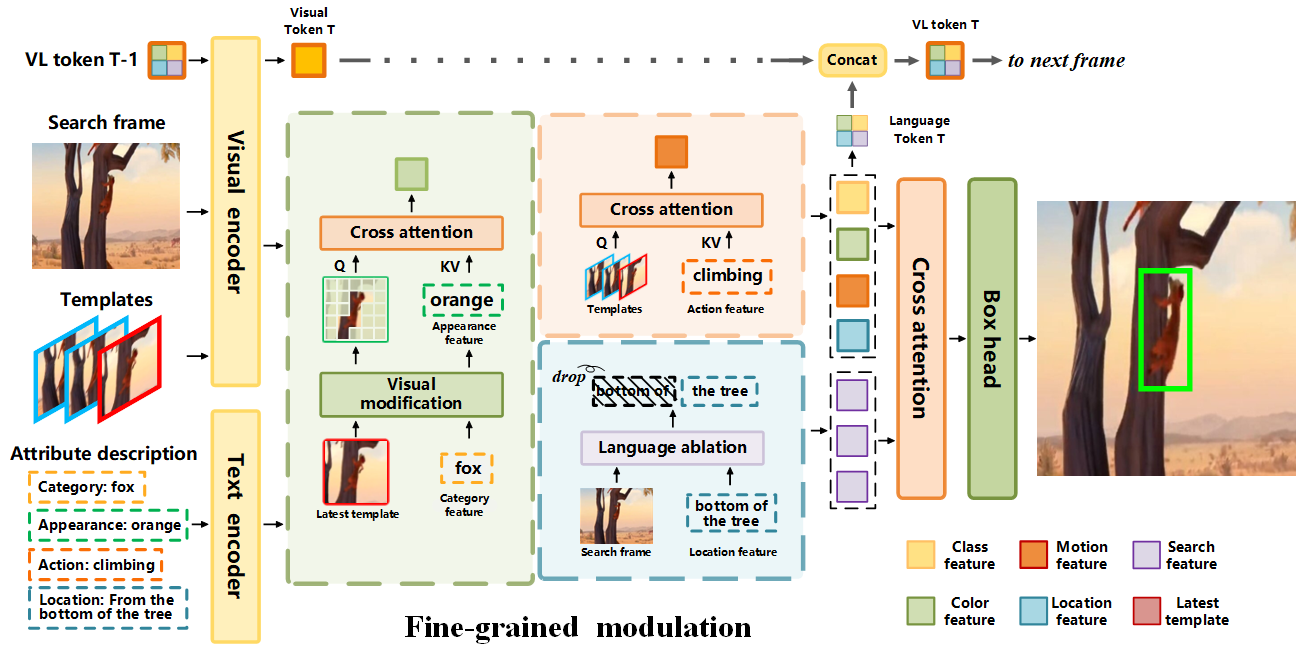}
    
   \caption{Overview of the proposed ATSTrack framework. ATSTrack has been improved in two aspects: 1) A Visual-language token is used to guide the extraction of visual features to obtain features that are more closely matched to the language description. 2) A Fine-Grained Modulation module is designed to make more effective modification to the language features}
   \label{fig:model} 
\end{figure*}
%-------------------------------------------------------------------------
\section{Related Work}
\subsection{Visual Single Object Trackers}
Single object tracking aims to locate the target in a video sequence according to the given bounding box in the first frame. 
Existing mainstream trackers~\cite{romtrack,Xie2022CorrelationAwareDT,siamcar,stark,hip,He2023TargetAwareTW,Kim2022TowardsST} typically rely on the matching between the template and the search region.
MixFormer~\cite{2mixformer} uses iterative mixed attention to integrate feature extraction and target information. 
OSTrack~\cite{ostrack} proposes a single stream framework that can jointly perform feature extraction and relation modeling and an early candidate elimination module to eliminate unnecessary search region tokens.

However, these methods may face significant challenge when the appearance of the target undergoes drastic changes (i.e., rapid motion or occlusion)\cite{Huang2024RTrackerRT}, since they use only the visual modality for feature relationship modeling,
Some methods have focused on utilizing motion information.
SeqTrack~\cite{seqtrack} models tracking as a sequence generation task, offers a simple framework by removing the redundant prediction head and loss function. 
ARTrack~\cite{art} treats tracking as a coordinate sequence interpretation task and uses a time autoregressive method to model changes in trajectory sequences, thereby maintaining cross-frame tracking of the target
Despite using additional motion information, these methods still heavily rely on visual matching and cannot completely eliminate the aforementioned limitation.

\subsection{Visual-Language Trackers}
 Visual-Language tracking aims to track targets based on visual features and additional natural language descriptions. since the rich semantic information in language description provides more accurate target reference.
 Li \etal~\cite{tnl} first introduces natural language into tracking achieving more robust results than visual tracker.
 The SNLT model~\cite{SNLT} uses language information and visual information to predict the state of the target individually and then fuses these predictions to obtain the final tracking result. 
 Guo \etal~\cite{divert} propose modality mixer for unified Visual-Language representation learning and the asymmetric searching strategy to mix Visual-Language representation.
 
 Recently, more researchers are beginning to notice the mismatch between visual inputs and language descriptions.
 Ma \etal~\cite{Decouple} decouple the tracking task into short-term context matching and long-term context perceiving to reduce the impact of misalignment.
 Shao \etal~\cite{querynlt} processes the inputs into prompts and proposes a multi-modal prompt modulation module to filter out prompts by leveraging the complementarity between visual inputs and language descriptions.
 Unlike other methods that rely on manual language annotations, CiteTracker~\cite{citetracker} uses CLIP~\cite{clip} to generate four initial attributes for the target and adjust the weights of these four attributes in each frame. 
 However, these methods still suffer from interference caused by the inherent difference in the temporal and spatial scale of information between visual and language features.
 To this end, we propose a novel framework which uses linguistic information to guide the extraction of visual features and modify language features in a fine-grained manner. 
\section{Method}
\subsection{Overview}
\cref{fig:model} shows the general framework of the TSATrack.
Unlike previous trackers, we replace the single template image with a sequence of templates to incorporate richer temporal information, while introducing an extra token as an additional input to the visual backbone. The output of visual backbone consists of: search feature $F_{\text{search}}$, template features $F_{\text{template}}=\left\{F_{\text{n}},F_\text{n-1},...,F_\text{init}\right\}$ and a visual token $T_{\text{vi}}$. 
We utilize a Large Language Model (LLM) to segment each language description into four phrases with different attributes based on their correspondence with visual inputs: \textbf{Category}, \textbf{Appearance}, \textbf{Action}, and \textbf{Location}. The language backbone subsequently extracts features of these various attributes: category feature $F_{\text{cate}}$, appearance feature $F_{\text{app}}$, action feature $F_{\text{act}}$ and location feature $F_{\text{loc}}$. 

These visual and language features are then fed into a Fine-Grained Modulation (FGM) module to acquire modified language features $F_{\text{lang}}=\left\{F_{\text{cate}},\overline{F}_{\text{app}},\overline{F}_{\text{act}},\overline{F}_{\text{loc}}\right\}$. We generate a language token $T_{\text{lang}}$ from modified language features $F_{\text{lang}}$ and aggregate $T_{\text{lang}}$ with $T_{\text{vi}}$ as the Visual-Language token $T_{\text{VL}}$, which is propagated to the visual backbone of the next frame to guide the extraction of visual feature. After that, $F_{\text{lang}}$ and the search feature $F_{\text{search}}$ are merged and send to the prediction head to obtain the tracking result.

\subsection{Visual Language Correspondence}
As previously mentioned, we segment each complete language description into four phrases with different attributes based on their correspondence with different visual inputs in terms of temporal and spatial scales: \textbf{Category}, \textbf{Appearance}, \textbf{Action}, and \textbf{Location}. For instance, ``Yellow airplane flying in the air" will be divided into \{``Category: airplane", ``Appearance: yellow", ``Action: flying", ``Location: in the air"\}, more examples are shown in \cref{fig:visual}. In this section, we provide a detailed explanation of these correspondences and the characteristics of different attributes.

\begin{figure}[t]
    \centering % 使用\centering来居中图片
    \includegraphics[width=0.9\linewidth]{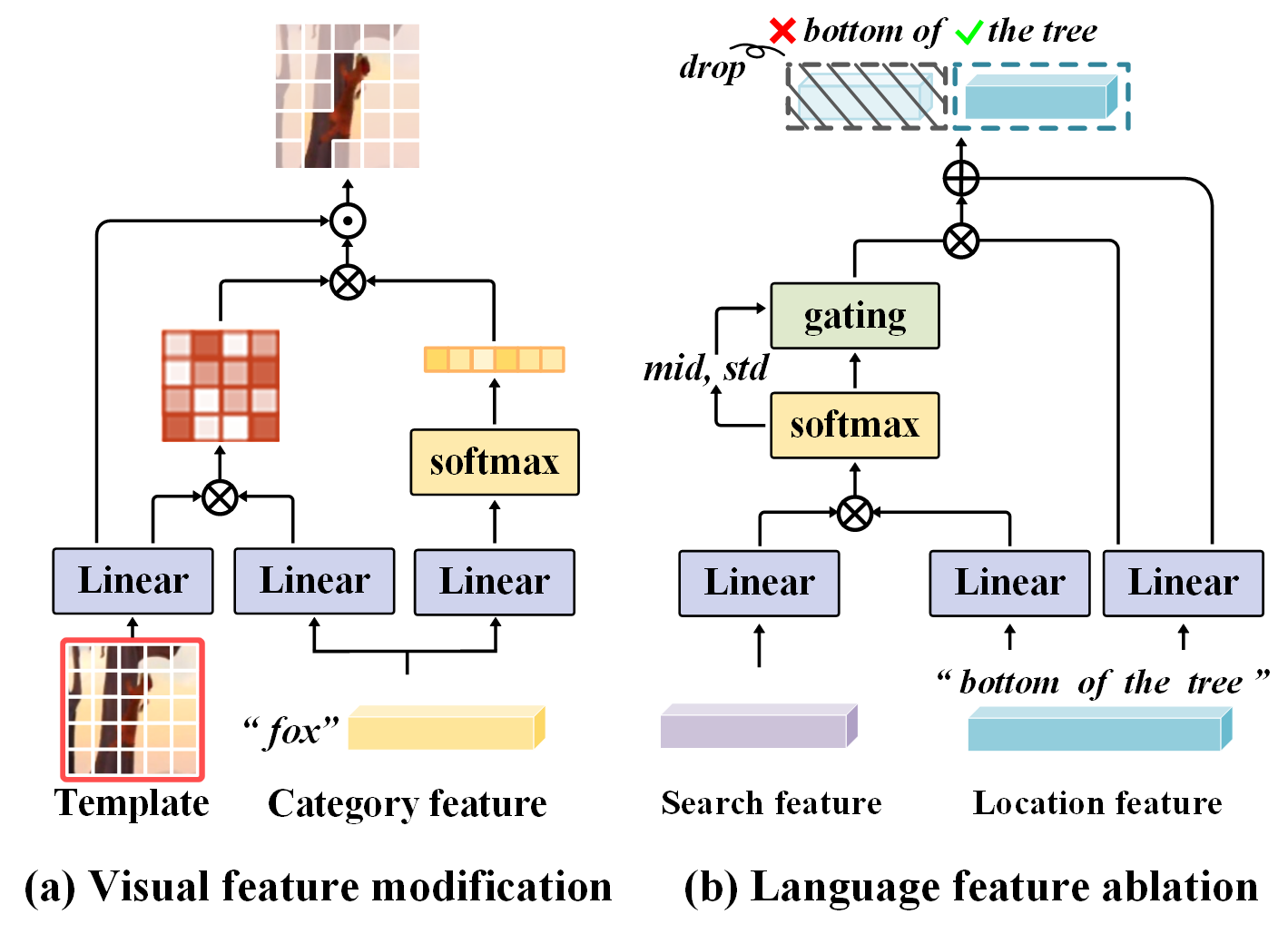}
     
   \caption{(a) The structure of the visual feature modification module. (b) The structure of the language feature ablation module.}
\label{fig:ablation}
\end{figure}
\noindent\textbf{Category and Appearance. } ``Category" and ``Appearance" correspond to the template from the latest frame rather than search frame, as template contains less background and can better reflect the object's category and appearance. The category descriptions are usually accurate and requires no further modification, while the appearance may vary, so we categorize them separately.

\noindent\textbf{Action. }``Action" refers to the motion state of the target. We consider ``Action" corresponds to the entire template sequence because it could be difficult to distinguish between actions such as ``walking" and ``running" using a single template. It should be noted that the interaction between the target and other objects is considered as 'location', as the other object may be far away from the target and thus not appear in the template.

\noindent\textbf{Location. }Descriptions of an object's Location often involve other objects in the background, so “Location” should correspond to the search image. As mentioned above, “Location” includes not only the literal description of where an object is located like, but also other descriptions that help locate the target, such as ``played by a man".

\subsection{Fine-Grained Modification}
The structure of the Fine-Grained Modulation (FGM) module are shown in~\cref{fig:model}. Compared to coarse-grained interaction used by previous trackers, fine-grained interaction can achieve better feature modification by manually aligning the temporal and spatial scales of different input components. Moreover, we have designed different modification strategies based on the unique characteristics of different inputs. As mentioned above, it is most ideal to modify appearance feature with template from the latest frame in the template sequence $F_{\text{n}}$. To prevent background from interfering with appearance feature, we employ a \textbf{Visual Feature Modification (VFM)}module that leverages $F_{\text{cate}}$ to suppress background information in $F_{\text{n}}$.
The action feature $F_{\text{act}}$ is modified by all template features $F_{\text{template}}=\left\{F_{\text{n}},F_\text{n-1},...,F_\text{init}\right\}$ through cross attention since they both contain temporal information. The mismatch between location descriptions and visual inputs is typically the most severe, with regard to this issue, we utilize a \textbf{Language Feature Ablation(LFA)} module to eliminate the mismatch parts in location features.

\noindent\textbf{Visual Feature Modification. }The purpose of Visual Feature Modification (VFM) is to suppress background information in the template features at pixel level. The structure of the VFM is illustrated in ~\cref{fig:ablation} (a). Given the category feature $F_{\text{cate}}$ and the template feature $F_{\text{n}}$ as input, we adopt linear projection layers to project them to same dimension and calculate the similarity matrix $M_{sim}$ between category and template features:
\[
M_{sim} = \text{softmax}\left(\frac{\delta_{t}(F_{\text{n}})\times \delta_{c}(F_{\text{cate}}) }{\sqrt{C}}\right)
\]
where $\delta_{c}$ and $\delta_{t}$  are projection layers for category features and template features. Since the importance of the information contained in different tokens of $F_{\text{cate}}$ also varies ~\cite{querynlt}, we calculate the importance score map and multiply it by $M_{sim}$ to increase the difference between target and background in the target map $M_t$. 
Finally, the modified template feature $\overline{F}_{\text{n}}$ is acquired by:
\[
M_t = M_{sim} \times  \text{softmax}\left(\delta_{t}(F_{\text{n}})\right)
\]
\[
\overline{F}_{\text{n}} = {F}_{\text{n}} \odot M_t
\]
The values in $M_t$ reflect the probability that the features belong to the target. Through this method, we can suppress the background features in the template and make a more accurate modification to the color features.

\noindent\textbf{Language Feature Ablation. }The core idea of Language Feature Ablation (LFA) is to filter out location information that are not align with the target's state, we achieve this by setting the aggregation weight of misaligned tokens to near $0$ through a gating operation.
The structure of the LFA is illustrated in ~\cref{fig:ablation} (b). In LFA, the similarity matrix $M_{sim}$ between search feature $F_{\text{search}}$ and location feature $F_{\text{loc}}$ is used as the weight to aggregate information in $F_{\text{loc}}$, the gating operation of $M_{sim}$ can be formulated as: 
\[
\theta = mid\left(M_{sim}^{j}\right) +\varphi std\left(M_{sim}^{j}\right)
\]
\[
G_j = \text{sigmoid}\left(\alpha \left(M_{sim}^{j}-\theta\right)\right)
\]
\[
M = M_{sim} \odot  G
\]
Where $\alpha=50$, $\varphi=0.5$. $M_{sim}^{j}$ is the $j_{th}$ column of $M_{sim}$. We use the weighted sum of the median and variance of $M_{sim}^{j}$ to initialize a threshold $\theta$, when the values in $M_{sim}^{j}$ are more discrete (\ie tokens in $F_{\text{loc}}$ have a greater difference in similarity), $\theta$ is also larger and has a better suppression effect. We subtract $\theta$ from $M_{sim}^{j}$ and multiply it with scaling factor $\alpha$ before applying the sigmoid function to obtain $G_j$, which represents the $j_{th}$ column of gating matrix $G$. The values in $G$ range from $0$ to $1$ and are directly proportional to the similarity scores in $M_{sim}$. By multiplying $G$ with $M_{sim}^{j}$, the weights of tokens in $F_{\text{loc}}$ that exhibit low similarity between $F_{\text{search}}$ will be projected to close to $0$. The modified location feature $\overline{F}_{\text{loc}}$ is acquired by:
\[
\overline{F}_{\text{loc}} = M \times \delta_{v}({F}_{\text{loc}})+{F}_{\text{loc}}
\]
where $\delta_{v}$  represents the projection layer for ${F}_{\text{loc}}$.

\subsection{Visual-Language Token}
Previous visual-language trackers methods usually confine the backbone's access to information to single modality, ignoring the need to extract features that are more relevant to the other modality. This overlook of cross-modality information interaction exacerbates the misalignment between visual and language features, thereby affecting the effectiveness of subsequent operations.

To address this issue, we generate a Visual-Language token $T_{\text{VL}}$ for each video frame and propagate it to the visual backbone of the following frame.
$T_{\text{VL}}$ is the aggregation of the visual token $T_{\text{vi}}$ and language token $T_{\text{lang}}$. Visual token $T_{\text{vi}}$ is the $cls$ token of the visual backbone, which consist of the global visual information. After acquiring the modified language features $F_{\text{lang}}$, we take the global average mean of $F_{\text{lang}}$ as language token $T_{\text{lang}}$ and concatenate $T_{\text{vi}}$ with $T_{\text{lang}}$ to acquire the Visual-Language token $T_{\text{VL}}$. The overall process can be formulated as:
 \[
T_{\text{lang}}=\text{avg}\left(\text{concat}\left[F_{\text{c}},\overline{F}_{\text{app}},\overline{F}_{\text{act}},\overline{F}_{\text{loc}}\right]\right)
\]
 \[
T_{\text{VL}}=\text{concat}\left[T_{\text{lang}} , T_{\text{vi}}\right]
\]
where concat[·,·] denotes the concatenation operation. 
 
$T_{\text{VL}}$ is concatenated with visual input of the next frame, by participating in subsequent attention operations within the visual backbone, $T_{\text{VL}}$ can serve as a guide for visual feature extraction. From the perspective of context understanding, $T_{\text{VL}}$ contains global visual and linguistic information from the previous frame, which helps the model to better model the temporal relationships between frames. From the perspective of Visual-Language alignment, the linguistic information contained in $T_{\text{VL}}$ guides the model to extract features that are more relevant to language descriptions.

\subsection{Prediction Head and Loss Function}
We employ a commonly used prediction head \cite{ostrack,Gao2023GeneralizedRM,10} comprising 3 conventional branches to obtain the center score map $C^{\frac{H_x}{p}\times \frac{H_x}{p}}$, an offset map $O^{2\times\frac{H_x}{p}\times \frac{H_x}{p}}$ and a normalized size
\begin{figure}[h]
    \centering % 使用\centering来居中图片
    \includegraphics[width=0.85\linewidth]{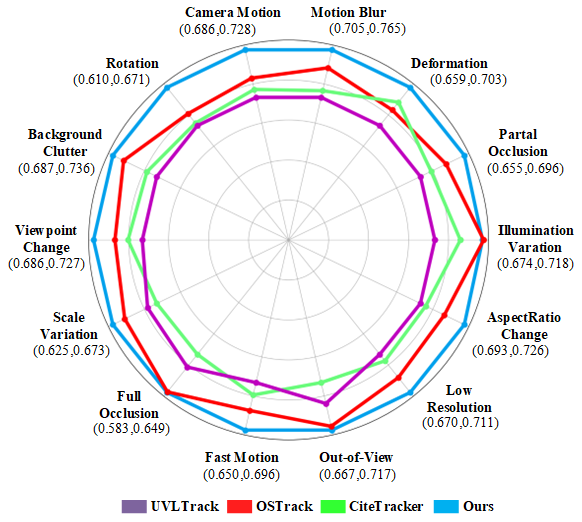}
     
   \caption{AUC score of different attributes in LaSOT.}
\label{fig:leida}
\end{figure}
map $S^{2\times\frac{H_x}{p}\times \frac{H_x}{p}}$, where p is the size of the image patches. The final tracking results are computed as follows:
\[
\left(x, y, w, h\right)=\text{map}\left(x_c+ O_x, y_c+O_y, S_x, S_y \right)
\]
where $\left(x_c, y_c\right)=\text{argmax}\left(C\right)$ and $\text{map}\left(\text{·}\right)$  represents the operation of mapping the bounding box back to its original size.

 We adopt the focal loss as classification loss $L_{cls}$, and the $L1$ loss and $GIoU$ loss.
 as regression loss. The overall loss function can be formulated as:
 \[
 L =L_{cls} +\lambda_1L_1 +\lambda_2L_{GIoU}
\]
We follow the setting in previous works and set $\lambda_1$ = 5 and $\lambda_2$ = 2 in our experiments.

\section{Experiment}
\subsection{Implementation Details}
The proposed model is implemented in Pytorch. The models are trained on 4 NVIDIA A6000 GPUs and tested on a single NVIDIA 3090 GPU.
We utilize the vanilla ViT-Base-384~\cite{VIT} pre-trained with MAE~\cite{MAE} as the visual backbone. The Clip-B-32~\cite{clip} model is selected as the language backbone.
We employ the AdamW to optimize the network parameters with initial learning rate of $1 \times10^5$ for the backbone, $1 \times10^4$ for the rest, and set the weight decay to $1 \times10^4$. We set the training epochs to 300 epochs with a batch size of 8. 60,000 image pairs are randomly sampled in each epoch. 

\begin{table*}
  \centering
  \begin{tabular}{l|c|ccc|ccc|ccc}
    \specialrule{1pt}{0pt}{0pt}
      \multirow{2}{*}{\textbf{Method}}&
      \multirow{2}{*}{\textbf{Source}}
      & \multicolumn{3}{c|}{\textbf{TNL2K}} 
      & \multicolumn{3}{c|}{\textbf{LaSOT}}
      & \multicolumn{3}{c}{$\mathbf{OTB_{lang}}$} \\
     \cline{3-11}
          && AUC & 
          $P_{norm}$ &
          P & 
          AUC & 
          $P_{norm}$ & 
          P &
          AUC & 
          $P_{norm}$ &
          P \\
    \hline
    \multicolumn{11}{c}{Visual trackers} \\
    \hline
    SwinTrack-B\cite{SwinTrack} &NIPS2022& 55.9 & - & 57.1 & 71.3 & - & 76.5 &- & - & - \\
    OSTrack\cite{ostrack}&ECCV2022 & 54.3 & - & - & 69.6  & 81.1 & 77.1 & - & - & - \\
    MixFormer-v2\cite{Cui2023MixFormerV2EF}&CVPR2022 & 57.4 & - & 58.4 & 70.6  & 80.8 & 76.2 & - & - & - \\
    ARTrack-B\cite{art} &CVPR2023& 58.9 & - & - & 72.6 & 81.7 & 79.1 & - & - & - \\
    SeqTrack-B\cite{seqtrack} &CVPR2023 & 56.4 & - & - & 71.5 & 81.1 & 77.8 &  - & - & - \\
    DropTrack\cite{drop} &CVPR2023 & 56.9 & - & 57.9 & 71.8 & 81.8 & 78.1 &  - & - & - \\
    AQATracker\cite{aqa} &CVPR2024& 59.3 & - & 62.3 & 72.7 & 82.9 & 80.2 & - & - & - \\
    ODTrack-B\cite{Zheng2024ODTrackOD} &AAAI2024& 60.9 & - & - & \textcolor{red}{73.2} & \textcolor{red}{83.2} & \textcolor{red}{80.6} & - & - & -  \\
    LoRAT-B\cite{lora} &ECCV2024& \textcolor{blue}{62.7} & - & \textcolor{blue}{63.7} &  \textcolor{blue}{72.9} &  \textcolor{blue}{81.9} &  \textcolor{blue}{79.1}& - & - & -  \\
         \rowcolor{gray!20}
     ATSTrack&Ours  & \textcolor{red}{66.2} & \textcolor{red}{84.2}  & \textcolor{red}{71.3} & 72.6  & 82.4 & 79.5 & \textcolor{red}{71.0} & \textcolor{red}{87.6} & \textcolor{red}{94.4} \\ 
    \hline
    \multicolumn{11}{c}{Visual-Language trackers} \\
    \hline
    SNLT\cite{SNLT} &CVPR2021& 27.6 & - & 41.9 & 54.0 & 63.6 & - & 66.6 & - & 80.4 \\
    VLT\cite{divert} &NIPS2022& 53.1 & - & 53.3 & 67.3 & - & 72.1& 65.3 & - & 85.6  \\
    JointNLT\cite{JointNLT}&CVPR2023 & 56.9 & 69.4 & 58.1 & 60.4 & 73.5 & 63.6 &65.3 & - & 85.6 \\
    DecoupleTNL\cite{Decouple}&ICCV2023 & 56.7 & - & 56.0 & \textcolor{blue}{71.2} & - & \textcolor{blue}{75.3} & \textcolor{red}{73.8} & - & \textcolor{red}{94.8} \\
    MMTrack\cite{MMTrack}&TCSVT2023 & 58.6 & 75.2 & 59.4 & 70.0 & 82.3 & 75.7 & 70.5 & - & 91.8 \\
    CiteTracker\cite{citetracker} &ICCV2023& 57.7 & 73.6 & 59.6 & 69.7 & 78.6 & 75.7 &  69.6 & 92.2 & 85.1 \\
    UVLTrack-B\cite{uvl}&AAAI2024& \textcolor{blue}{63.1} & - & \textcolor{blue}{66.7} & 69.4 & - & 74.9 & 69.3 & - & 89.9 \\
    QueryNLT\cite{querynlt}&CVPR2024 & 57.8 & 75.6 & 58.7 & 59.9 & 69.6 & 63.5 & 66.7 & 82.4 & 88.2  \\
         \rowcolor{gray!20}
     ATSTrack&Ours  & \textcolor{red}{66.2} & \textcolor{red}{84.2} & \textcolor{red}{71.3} & \textcolor{red}{72.6}  & \textcolor{red}{82.4} & \textcolor{red}{79.5} & \textcolor{blue}{71.0} & \textcolor{blue}{87.6} & \textcolor{blue}{94.4} \\
    \hline
  \end{tabular}
  \caption{Comparison with both state-of-the-art visual and visual-language trackers on TNL2K, LaSOT, $\text{LaSOT}_{ext}$, and $\text{OTB}_{lang}$. The best two results in each parts are shown in \textcolor{red}{red} and \textcolor{blue}{blue} respectively.}
  \label{tab:sota}
\end{table*}
Our training dataset comprises TNL2K~\cite{tnl2k}, LaSOT ~\cite{lasot} GOT-10k~\cite{got10k} and TrackingNet~\cite{trackingnet}, with an equal sampling ratio across the datasets. TNL2K and LaSOT contain manually annotated language descriptions, we use LLM to segment the language descriptions into different attributes. GOT-10k includes annotations for category and motion, and we set other attributes to ``None". TrackingNet contains category labels, we use the pre-trained Clip model in ~\cite{citetracker} to predict the color of each target. 

\subsection{State-of-the-art Comparison}
We compare our tracker with both state-of-the-art visual and visual-language methods on three commonly used datasets with language annotation, including TNL2K, LaSOT, and $\text{OTB}_{lang}$.
Results are shown in Table. \ref{tab:sota}.

\noindent\textbf{TNL2k} ~\cite{tnl2k} is a benchmark specifically dedicated to the tracking-by-language task, which contains a total of 2k sequences and 663 words. The benchmark introduces two new challenges, i.e.adversarial samples and camera switching, which makes it a robust benchmark.
Our method demonstrates substantial performance enhancement on the TNL2k benchmark. Specifically, the proposed ATSTrack report an AUC of $\text{66.2}\%$ and surpass state-of-the-art visual and visual-language trackers by $\text{3.5}\%$ and $\text{3.1}\%$ respectively. The favorable performance demonstrates the promising potential of our tracker to deal with adversarial samples and modality switch problems.

\noindent\textbf{LaSOT }~\cite{lasot} is a large-scale long-term tracking benchmark with an average video length of more than 2,500 frames. It includes 1120 sequences for training and 280 sequences for testing. ATSTrack outperforms the second best visual-language tracker by $\text{1.8}\%$ in term of AUC, meanwhile achieves a performance comparable to SoTA visual trackers. Furthermore, ~\cref{fig:leida} shows the detailed results on different attributes in LaSOT. Our model outperforms other tracking methods on multiple challenge attributes. These result shows that ATSTrack could better utilizes information from both modalities compared to other visual-language trackers and have superior long-term tracking capabilities.

\noindent\textbf{$\text{OTB}_{lang}$ }~\cite{SNLT} is OTB-100~\cite{OTB} dataset extended with a language description of the target object per sequence. It encompasses 11 challenging interference attributes, such as motion blur, scale variation, occlusion, and background clutter. ATSTrack achieves the second best performance with an AUC of $\text{71.0}\%$ and precision of $\text{94.4}\%$, surpassing the third best tracker by $\text{2.6}\%$ in terms of precision.

\subsection{Ablation Study}
We conduct ablation studies on the LaSOT dataset to verify the effectiveness of each component in our model.

 \begin{table*}
  \centering
\begin{subtable}[t]{0.33\textwidth}
% \small
\footnotesize
  \centering
  \begin{tabular}{l|ccc}
        \specialrule{1pt}{0pt}{0pt}
          Method
              & AUC & 
              $P_{norm}$ &
              P 
            \\
        \hline
        Baseline & 70.6 & 80.7 & 77.1 \\
        w/o FGM & 71.1 & 80.6 & 77.4 \\
        w/o VFM & 71.6 & 81.5 & 78.4\\
        w/o LFA & 71.5 & 81.2 & 78.4\\
        w/ FGM  & \textcolor{red}{72.0} & \textcolor{red}{82.1} & \textcolor{red}{79.0} \\
        \hline
      \end{tabular}
      \caption{Ablation study of the Fine-Grained Modulation.}
      \label{tab:ablation1}
\end{subtable}
  \hfill
\begin{subtable}[t]{0.33\textwidth}
% \small
\footnotesize
  \centering
\begin{tabular}{c|ccc}
        \specialrule{1pt}{0pt}{0pt}
              Method
              & AUC & 
              $P_{norm}$ &
              P  
              \\
        \hline
        w/o token  & 72.0 & 82.1 & 79.0\\
        w/o V token & 71.7 & 82.0 & 78.6 \\
        w/o L token & 72.0 & 82.4 & 78.6\\
        Attn & 72.4 & \textcolor{red}{82.6} & \textcolor{red}{78.9}\\
        Concat & \textcolor{red}{72.6} & 82.4 &79.5\\
        \hline
      \end{tabular}
      \caption{Ablation study of the Visual-Language token. }
      \label{tab:ablation2}
\end{subtable}
  \hfill
\begin{subtable}[t]{0.33\textwidth}
  \centering
  % \small
  \footnotesize
  \begin{tabular}{c|ccc}
    \specialrule{1pt}{0pt}{0pt}
          \textbf{\textbf{$\alpha$}}
          & AUC & 
          $P_{\text{norm}}$ &
          P  \\
    \hline
    500 & 71.4 & 81.0 & 77.8\\
    100 & 71.9 & 81.7 & 78.7 \\
    50& \textcolor{red}{72.0} & \textcolor{red}{82.1} & \textcolor{red}{79.0} \\
    25 & 71.3 & 81.2 & 78.0 \\
    \hline
  \end{tabular}
  \caption{Comparison of different gating weight in LFA.}
  \label{tab:a}
\end{subtable}
  \caption{Ablation Studies of  modules in ATSTrack. The best result are shown in \textcolor{red}{red}}
  \label{tab:study}
\end{table*}

\noindent\textbf{Effect of Fine-Grained Modulation.}
The ablation results of FGM are shown in Table. \ref{tab:ablation1}. We construct a \textbf{baseline} by removing components related to language and token propagation mechanism from our model. Performing coarse-grained interaction between language features and visual features through cross-attention (\textbf{w/o FGM}) leads to an increase in the AUC score by $\text{0.5\%}$ on LaSOT, demonstrating the advantage of using language descriptions in tracking task. \textbf{w/ FGM} shows that the use of fine-grained modulation improved the AUC score by and $\text{1.4\%}$ compared to the baseline, demonstrating the necessity of reduce the affect caused by the temporal and spatial difference between modality.
We also verify the effectiveness of Visual Feature Modification (VFM) module and Language Feature Ablation (LFA) module by replacing them with regular cross attention. The results show that \textbf{VFM} improves the AUC score by $\text{0.4\%}$, and the \textbf{LFA} improves the AUC score by $\text{0.6\%}$.

\begin{table}
  \centering
\adjustbox{width=0.51\linewidth}{
      \begin{tabular}{c|ccc}
    \specialrule{1pt}{0pt}{0pt}
          Attr
          & AUC & 
          $P_{norm}$ &
          P  \\
    \hline
    w/o Cate
     & 72.3 & 82.2 & 78.7 \\
    w/o App & 72.1 & 81.8 & 78.8 \\
    w/o Act & 72.5 & 82.1 & 79.2 \\
    w/o Loc & 72.4 & 82.6 & 79.1 \\
    Full & \textcolor{red}{72.6} & \textcolor{red}{82.4} & \textcolor{red}{79.5} \\
    \hline
  \end{tabular}
  }
      \caption{Effect of different attribute descriptions. The bset results are marked in \textcolor{red}{red}.}
      \label{tab:ablation3}
\end{table}

\begin{figure}[t]
    \centering % 使用\centering来居中图片
    \includegraphics[width=0.9\linewidth]{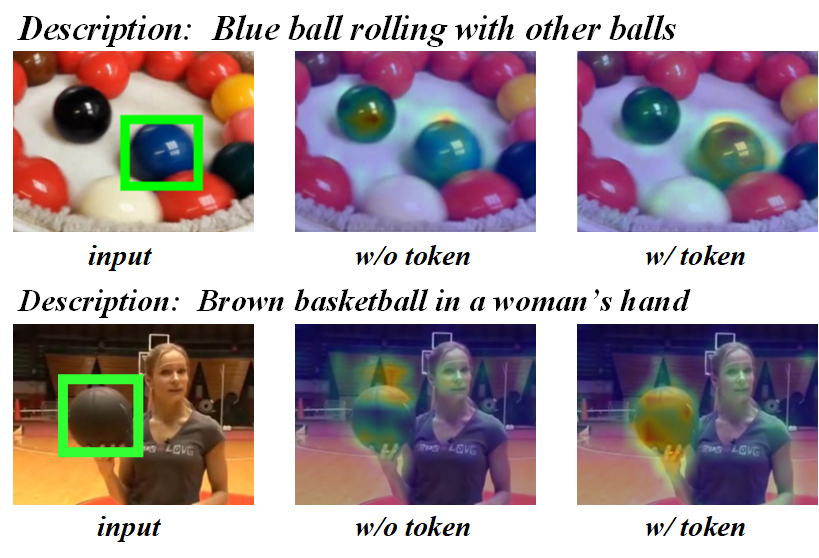}
     \vspace{-10pt}
   \caption{The attention map of Visual-Language token.}
\label{fig:heatmap}
\vspace{-10pt}
\end{figure}

\noindent\textbf{Gating Weight in LFA}
We analyze the impact of different gating weights $\alpha$ on the effect of different on LFA performance on LaSOT. As shown in \cref{tab:a}, LFA achieves the best performance with $\alpha=25$. Since the attention matrix is normalized, the disparity between each attention score and the threshold remains relatively minor. When $\alpha$ is small, the sigmoid function will be too smooth, leading to an indistinct difference between tokens of different attention scores. When $\alpha$ is large, the disparity between tokens that exceed and below the threshold becomes too large, and the difference within their respective classes will be insignificant.

\noindent\textbf{Effect of Visual-language token.} The ablation results of FGM are shown in ~\cref{tab:ablation2}. Without using the Visual-language token (\textbf{w/o token}), the model decreases in the AUC score by $\text{0.6\%}$. This validates the effectiveness of the Visual-language token. We further analyze the influence of information from different modalities. Using \textbf{the visual token independently (w/o L token)} does not leads to notable changes, as visual tokens only encompass global visual information and could not bridge the gap between visual and language modalities. Using
\textbf{the Language token independently (w/o V token)} leads to a decrease in the AUC score by $\text{0.3\%}$, the reason could be the semantic level misalignment between language and visual features. These results show that both global visual features and language features are essential to help the model better understand the target features. We compare different ways to aggregate visual and language information. We have found that performing cross attention between tokens slightly improves the precision but leads to AUC decrease compared to concatenation and chose to concatenate visual and language tokens to acquire VL token in our model.

\noindent\textbf{Effect of Each attribute.} An important issue in visual-language tracking lies in determining which kind of descriptions are most conducive to effective tracking. Given that we have segmented language descriptions into different attributes, it becomes convenient for us to perform ablation studies on them.
As shown in ~\cref{tab:ablation3}, Removing \textbf{category descriptions (w/o Cate}) leads to a decrease in AUC by $\text{0.3\%}$, demonstrating the effect of category descriptions.
Removing \textbf{appearance descriptions (w/o App}) causes a notable decrease in AUC by $\text{0.5\%}$, as appearance is usually the most obvious factor to distinguish the target from other objects. It should be noted that since existing datasets provide fewer appearance descriptions compared with other attributes, its actual effect would be greater. 
\begin{figure*}[t]
\vspace{10pt}
    \centering % 使用\centering来居中图片
    \includegraphics[width=0.9\linewidth]{
    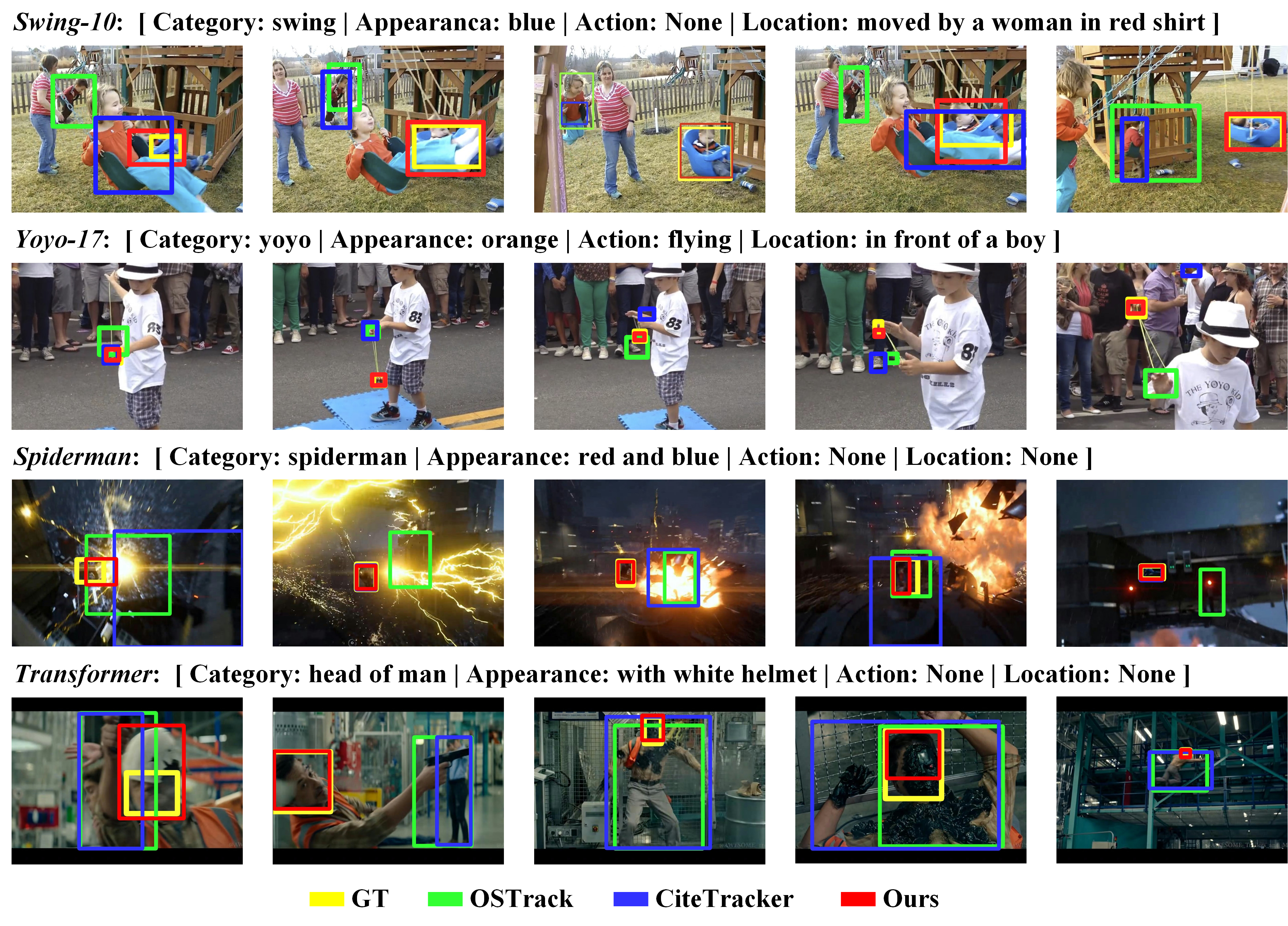}
    
   \caption{ Visualized results of the proposed ATSTrack on three challenging scenarios for visual object tracking: severe occlusion, fast motion and view change. Results show that ATSTrack outperforms other advanced trackers on these challenging sequences.}
   \label{fig:visual} 
\end{figure*}
\textbf{The action descriptions (w/o act}) has the weakest impact on tracking results. We consider the reason that action is only useful to distinguishing targets from other similar objects. However, similar objects always share the same actions in existing datasets. 
\textbf{Location descriptions (w/o loc}) also has a weak affect on tracking result. Consider that the location feature are already modified by the LFA module, we believe existing location description are more likely to cause interference rather than enhance tracking. 

\subsection{Visualization}
To intuitively demonstrate the excellent performance of the proposed method, we visualize the tracking results of our model and two advanced trackers: OSTrack\cite{ostrack} and CiteTracker\cite{citetracker}. In ~\cref{fig:visual}, the challenge of performing visual tracking on these four sequences arises from severe occlusion (Swing, Spiderman), fast motion (Yoyo, Spiderman), and view changes (Transform). In contrast, the language descriptions offer accurate information about the target and could be leveraged to achieve more robust tracking. The results show that our proposed ATSTrack outperforms other trackers in these three scenarios, indicating its ability to fully utilize advanced semantic information contained in language descriptions.

Furthermore, we visualize the change of attention maps after introducing the Visual-Language token. As shown in ~\cref{fig:heatmap}, in the ball sequence, the visual backbone pays more attention to the target than distracting object (blackball). In the basketball sequence, the model pays more attention to important elements referenced in the language description (basketball and woman) and reduces the focus on irrelevant texture in the background. These results indicate that the Visual-Language token meets our expectation of guiding the model to extract visual features that are more aligned with language descriptions.

\subsection{Conclusion}
In this work, we present ATSTrack, which enhances the effect of visual-language tracking by obtaining features with better alignment. Specifically, we segment language descriptions into different attributes based on their temporal and spatial correspondence with visual inputs, and modify their features in a fine-grained manner, thereby reducing the interference caused by the difference in the temporal and spatial scale of information between visual and language modality. Moreover, we introduce a Visual-Language token that comprises modified linguistic information from the previous frame to guide the model to extract visual features that are more relevant to language description.  Extensive experiments show that ATSTrack can effectively use the information from visual and language modality and achieves a performance comparable to existing methods.

\clearpage
{
    \small
    \bibliographystyle{ieeenat_fullname}
    \bibliography{main}
}

\end{document}